\pdfoutput=1

\documentclass[11pt]{article}

\usepackage[preprint]{acl}

\usepackage{times}
\usepackage{latexsym}
\usepackage{afterpage}

\usepackage[T1]{fontenc}

\usepackage[utf8]{inputenc}

\usepackage{microtype}

\usepackage{inconsolata}

\usepackage{graphicx}

\usepackage{multirow}
\usepackage{enumitem}
\usepackage{xcolor,colortbl}
\usepackage{soul}
\sethlcolor{yellow}
\usepackage{tabularx}
\usepackage{tcolorbox}
\usepackage{comment}
\usepackage{amsmath}  
\usepackage{booktabs}
\usepackage{breqn}
\usepackage{soul}
\usepackage{xcolor,colortbl} 
\usepackage{changepage,threeparttable} 
\usepackage{adjustbox}
\usepackage{lineno}
\usepackage{makecell}
\usepackage{url}
\title{Modeling Professionalism in Expert Questioning \\ through Linguistic Differentiation}

\author{Giulia D'Agostino \\
  Università della Svizzera italiana, \\ Switzerland \\
  \texttt{dagosgi@usi.ch} \\\And
  Chung-Chi Chen \\
  National Institute of Advanced \\ Industrial Science and Technology, \\ Japan \\
  \texttt{c.c.chen@acm.org} \\}

\begin{document}
\maketitle

\begin{abstract}
Professionalism is a crucial yet underexplored dimension of expert communication, particularly in high-stakes domains like finance. This paper investigates how linguistic features can be leveraged to model and evaluate professionalism in expert questioning. We introduce a novel annotation framework to quantify structural and pragmatic elements in financial analyst questions, such as discourse regulators, prefaces, and request types. Using both human-authored and large language model (LLM)-generated questions, we construct two datasets: one annotated for perceived professionalism and one labeled by question origin. We show that the same linguistic features correlate strongly with both human judgments and authorship origin, suggesting a shared stylistic foundation. Furthermore, a classifier trained solely on these interpretable features outperforms gemini-2.0 and SVM baselines in distinguishing expert-authored questions. Our findings demonstrate that professionalism is a learnable, domain-general construct that can be captured through linguistically grounded modeling.
\end{abstract}

\section{Introduction}
Professionalism is a vital yet underdefined dimension of language use in high-stakes communicative settings. Whether in corporate finance, medical consultations, legal interviews, or hiring processes, expert questioning demands more than fluency or grammatical correctness—it requires clarity, relevance, and strategic pragmatics. Despite its intuitive importance, modeling professionalism in language remains elusive, in part due to the lack of consistent, operational definitions and annotated datasets.
Current NLP systems excel at generating well-formed text but often struggle with the subtleties of professional discourse. In expert interactions, subtle linguistic cues, such as how a question is framed, how background is contextualized, or how politeness is conveyed, can determine whether an utterance is perceived as informed and effective or vague and unprofessional. These micro-level elements are difficult to define explicitly, but they are deeply rooted in linguistic structure.

We argue that such linguistically grounded features offer a promising path forward. Rather than relying solely on surface metrics like sentence length or lexical complexity, we focus on structural and pragmatic properties of questions: the use of prefaces, the types of requests embedded, the presence of discourse regulators, and more. These elements reflect not just what is being asked, but how the speaker positions themselves in relation to the audience, the topic, and the communicative norms of the domain.
In this work, we explore whether such features can be operationalized to distinguish between more and less professional forms of expert questioning. We develop an annotation pipeline to extract and quantify a range of linguistically motivated features, and we investigate how these features correlate with human perceptions of professionalism.

To evaluate their predictive power, we introduce a controlled classification task. Specifically, we examine whether questions authored by human experts can be reliably distinguished from those generated by large language models (LLMs), under matched conditions. While our ultimate goal is to model professionalism, this origin-based contrast serves as a proxy—allowing us to isolate the stylistic and pragmatic signatures of expert discourse in a structured, observable way.
Our results show that linguistic features not only support accurate origin classification, but also align strongly with human judgments of professionalism. This provides both a practical tool for evaluating generated language and a theoretical foundation for future work on modeling professionalism in a domain-general, interpretable manner.

\section{Related Work}

Professional communication differs from everyday language through its use of formal, goal-oriented, and domain-specific features. Prior work shows that institutional discourse is shaped by roles and communicative goals, producing distinctive linguistic patterns \citep{DrewHeritage1992}. Across domains like medicine and law, professionals use technical terms and polite, precise language to convey expertise \citep{SarangiRoberts1999}. These features serve functional roles—ensuring clarity, maintaining face, and aligning with institutional norms. Our work builds on this foundation by treating professionalism as a measurable linguistic trait in expert communication.

Questions in expert settings often serve strategic purposes beyond information seeking. Studies highlight how their design—yes/no vs. wh-, direct vs. indirect—can signal stance and control responses \citep{ClaymanHeritage2002}. For example, adversarial questions can exert pressure or challenge authority \citep{ClaymanEtAl2007}. Professionals across domains similarly embed rhetorical aims in question form, balancing politeness with critical inquiry \citep{Ilie2021}. Our approach draws on these findings to identify linguistic markers of professional questioning.

LLM outputs differ systematically from human writing, even when fluent. Corpus studies show LLMs tend to produce more uniform, less varied, and less contextually rich language \citep{Sardinha2024, ZanottoAroyehun2024}. These differences inform detection methods and motivate our work: we focus on the nuanced features of professional human questioning that current models struggle to replicate. Rather than pure classification, we analyze interpretable linguistic indicators of professionalism in human vs. AI-generated questions.

Earnings calls are a prime domain for studying expert dialogue. These events feature formal yet interactive Q\&A between executives and analysts, marked by strategic questioning within professional constraints \citep{Camiciottoli2010Earnings, OliveiraPereira2018}. Analysts often use indirect or framed questions to press for clarity while maintaining collegiality. Language choices in this context can impact perceptions and market outcomes. Our work extends prior studies by computationally modeling professionalism in analyst questions, using human-AI differences as a lens to capture these expert traits \citep{palmieri_argumentation_2015, dagostino_capturing_2024}.

\section{Connecting Professionalism and Question Origin}

In this section, we explore the relationship between perceived professionalism and question origin. While our ultimate goal is to model professionalism as it is perceived by human evaluators, directly learning such a construct is challenging due to the subjectivity and variability of human judgments. Instead, we propose that comparing questions authored by domain experts to those generated by large language models (LLMs) offers a viable contrastive setting that surfaces the linguistic cues associated with professional communication.

We present three empirical components supporting this claim. First, we introduce the datasets used in our study, including a crowdsourced annotation of professionalism scores. Second, we describe the linguistically motivated features used to characterize each question. Finally, we analyze how these features correlate with both human professionalism ratings and question origin, showing that the same stylistic and pragmatic traits are salient in both tasks.

\subsection{Datasets and Annotation}

To investigate professionalism in expert questioning, we construct two primary datasets. The first, which we refer to as the Human-Rated Professionalism Dataset (HRPD), consists of 250 questions sampled from earnings call Q\&A sessions, equally balanced between those asked by financial analysts and those generated by LLMs. Each question was rated by five independent annotators on a 3-point scale for perceived professionalism in previous work by~\citet{juan_co-trained_2024}.\footnote{\citet{juan_co-trained_2024} shared the dataset upon request.} This dataset provides a distribution of subjective human judgments, enabling us to examine which linguistic features align with perceived professionalism.

The second dataset, which we term the Question Origin Dataset (QOD), combines a separate set of 250 analyst questions and 250 LLM-generated questions, sampled from a broader pool but without human rating. This dataset is used for training and evaluating classification models on the binary task of determining whether a question is of human or machine origin. While the labels in this dataset reflect authorship, not professionalism per se, we treat this task as a structured proxy to study how professional linguistic traits manifest at scale.

\subsection{Linguistic Features}

In order to characterize the pragmatic and structural properties of questions within financial dialogue, we rely on a linguistically-motivated feature set. These features are selected for their theoretical grounding in discourse analysis, pragmatics, and question typology, and they allow us to move beyond surface-level NLP metrics to a more interpretable, fine-grained representation of question design. We divide them into four principal categories: discourse regulators, prefaces, question types, and request types.

Discourse regulators are dialogical or meta-discursive components that organize the interactional flow of the question turn. They may acknowledge the previous speaker, specify who the question is for, indicate the number or order of questions, or introduce the topic explicitly. For example, expressions such as ``Thanks,'' ``Chris, this one is for you,'' or ``A quick one on margins'' serve to mediate the exchange in ways that are socially and structurally relevant. These elements are key in professional financial exchanges, where clarity, politeness, and sequential control are valued.

Prefaces, by contrast, are propositions that precede the question proper and function to justify, contextualize, or rationalize the upcoming interrogative act. They may state facts (``You said margins were up last quarter''), express opinions (``It seems like guidance is too optimistic''), report others' speech (``As the CEO mentioned in the last call...''), or include meta-commentary (``I ask this because it wasn’t clear from the disclosure''). These reveal the epistemic stance of the speaker and are frequently used in professional contexts to construct credibility or soften face-threatening acts.

We also categorize the syntactic form of the question into three types: open (wh-questions), polar (yes/no), and closed-list (multiple-choice). These forms determine the amount and specificity of information requested and reflect the speaker’s strategic intent. Finally, we annotate the request type encoded in the question, that is, the kind of answer being sought—ranging from clarification and confirmation to explanations, data, or opinions. This level of pragmatic annotation allows us to approximate the functional load of each question.

To determine how these linguistic features align with perceived professionalism, we compute Spearman correlations between each feature and the professionalism scores gathered via questionnaire. Table~\ref{tab:features-and-professionalism-correlation-reordered} presents all statistically significant correlations in both the \textit{mixed} and \textit{combined} datasets, with positively correlated features listed first (top-down) and negatively correlated ones following.

\begin{table}[t]
    \centering
    \begin{tabular}{l|rr}
        & \texttt{HRPD} & \texttt{QOD} \\
        \hline
        \textbf{Request types} & & \\
        \quad \textit{explanation} & ↑↑↑ & ↑ \\
        \quad \textit{clarification} & ↓↓↓ & ↓↓ \\
        \quad \textit{confirmation} & ↓↓↓ & ↓↓ \\
        \hline
        \textbf{Discourse regulators} & & \\
        \quad \textit{acknowledgment} & ↑↑↑ & ↓↓↓ \\
        \quad \textit{recipient} & ↓↓↓ & ↓ \\
        \quad \textit{theme} & ↓↓↓ & ↓↓↓ \\
        \quad \textit{enumeration} & ↓↓↓ & ↓↓↓ \\
        \quad \textit{counting} & ↓↓↓ & ↓↓↓ \\
        \quad \textit{inside comment} & ↓↓↓ & ↓↓↓ \\
        \hline
        \textbf{Prefaces} & & \\
        \quad \textit{reported speech} & ↓↓ & ↓↓↓ \\
        \quad \textit{opinion} & ↓↓↓ & ↓↓ \\
        \quad \textit{fact} & ↓↓↓ & ↓↓↓ \\
        \quad number & ↓↓↓ & ↓↓↓ \\
        \quad length & ↓↓↓ & ↓↓↓ \\
        \hline
        \textbf{Question types} & & \\
        \quad \textit{open} & ↓↓↓ & ↑ \\
        \quad \textit{polar} & ↓↓↓ & ↓↓↓ \\
        \quad \textit{closed-list} & ↓↓↓ & ↓↓ \\
        \hline
        \textbf{NLP features} & & \\
        \quad type-token ratio & ↑↑↑ & ↑↑↑ \\
        \quad Flesch-Kincaid & ↑↑↑ & ↑↑↑ \\
        \quad Dale-Chall & ↑↑↑ & ↑↑↑ \\
        \quad word count & ↓↓↓ & ↓↓↓ \\
        \quad sentence count & ↓↓↓ & ↓↓↓ \\
        \quad NER (person) count & ↓↓↓ & ↓↓ \\
        \quad stopword count & ↓↓↓ & ↓↓↓ \\
    \end{tabular}
    \caption{Spearman correlations with perceived professionalism. Arrows indicate direction and strength: ↑ = p < .05, ↑↑ = p < .01, ↑↑↑ = p < .001 (↓ for negative).}
    \label{tab:features-and-professionalism-correlation-reordered}
\end{table}

\subsection{Converging Correlates}

To assess whether perceived professionalism and question origin share common linguistic foundations, we compute Spearman correlations between the annotated features and two target variables:
\begin{enumerate}
    \item \textbf{Perceived Professionalism:} The mean human rating (1–3) from the Human-Rated Professionalism Dataset.
    \item \textbf{Question Origin:} A binary label (human or LLM-generated) from the Question Origin Dataset.
\end{enumerate}

Table~\ref{tab:features-and-professionalism-correlation-reordered} presents the features that are significantly correlated with both variables across the two datasets. We observe that many of the same linguistic indicators—such as preface frequency and length, request and question types, discourse structure, and readability—exhibit consistent correlation direction and strength in both tasks. For example, longer or more numerous prefaces correlate negatively with professionalism ratings and are more prevalent in machine-generated questions. Conversely, questions with clearer thematic focus and higher readability scores tend to be both rated as more professional and identified as human-authored.
This convergence suggests that the linguistic traits perceived by humans as hallmarks of professionalism are the same traits that distinguish human expert questioning from LLM-generated imitations. We interpret this alignment as evidence that the question origin task can serve as a valid, linguistically grounded proxy for modeling professionalism, particularly in the absence of reliable large-scale annotation.

Interestingly, our results also challenge the common intuition that professional language must be complex or dense. In contrast, we find that higher professionalism ratings are associated with greater readability, fewer words and sentences, and more concise structure overall. This suggests that perceived professionalism may align more strongly with clarity and rhetorical control than with verbosity or technical elaboration.
Finally, we observe higher variability in human ratings for naturally occurring questions, possibly due to their syntactic or pragmatic complexity. This further supports our use of question origin as a stable proxy for initial modeling, while acknowledging that true professionalism is inherently gradient and context-sensitive.

\section{Predicting Professionalism with Linguistic Features}

To evaluate whether linguistic features alone can predict professionalism, we train a Random Forest classifier using the full set of annotated features identified in earlier sections. We compare this model against two baselines:

\begin{itemize}
    \item \textbf{SVM:} A support vector machine trained on raw question text using TF-IDF features.
    \item \textbf{gemini-2.0-flash (few-shot):} A prompt-based classifier using in-context examples, asked to predict whether a question was authored by a human or generated by a model. 
\end{itemize}

Table~\ref{tab:final-prediction-results} reports accuracy and $F_1$ scores for the full classification task. The Random Forest model, despite using no text input, outperforms both baselines.

\begin{table}[t]
    \centering
    \begin{tabular}{l|cc}
        \textbf{Model} & \textbf{Accuracy} & \textbf{F\textsubscript{1}} \\
        \hline
        Gemini & 0.89 & 0.89 \\
        SVM & 0.92 & 0.92 \\
        Random Forest & \textbf{0.96} & \textbf{0.96} \\
    \end{tabular}
    \caption{Classification accuracy and $F_1$ score on the full question set. Our feature-based model outperforms both the SVM baseline and gemini-2.0-flash.}
    \label{tab:final-prediction-results}
\end{table}

These results offer two key insights. First, the linguistic features identified in earlier analysis are not only correlated with professionalism, but are also predictive. Their discriminative power is sufficient to classify questions as human- or model-authored with high accuracy, surpassing even a large, general-purpose language model using in-context learning.
Second, the fact that a shallow model using explicitly defined, interpretable features outperforms a black-box LLM suggests that professionalism is not an emergent property requiring deep modeling, but rather one that is encoded in recognizable and theoretically grounded linguistic signals. This supports our overall argument: that linguistic theory can guide effective, transparent modeling of communicative quality, especially in domains where clarity, etiquette, and rhetorical function matter.

\section{Conclusion}

We present a linguistically grounded approach to modeling professionalism in expert questioning. By analyzing structural and pragmatic features, we show that the same cues align with both human judgments and question origin. Our findings challenge the notion that complexity signals professionalism, and show that concise, readable questions are rated more professional. A lightweight classifier using these features outperforms both SVM and gemini-2.0-flash, demonstrating that professionalism is not only learnable, but also expressible through interpretable linguistic signals.

\section*{Limitations}

Our study focuses on a single domain—financial earnings calls—which may limit generalizability to other professional contexts such as law or healthcare. While we treat question origin as a proxy for professionalism, this distinction cannot fully capture the gradient and context-sensitive nature of the construct. Moreover, although our crowdsourced annotations offer useful insights, variability in human ratings suggests that perceptions of professionalism are not uniform and may depend on factors beyond linguistic form.

\section*{Supplementary materials availability statement}
The Question Origin Dataset and the test dataset for prediction evaluation are available at \url{https://github.com/dagosgi/question-origin_professionalism_short/tree/main/supplementary-materials}.

\bibliography{ref}

\appendix

\section{All Linguistic Features}
Table~\ref{tab:all features} shows all features we used in the experiments.

\begin{table}[t]
    \centering
    \begin{tabular}{l|rr}
        & \texttt{HRPD} & \texttt{QOD} \\
        \hline
        \textbf{Request types} & & \\
        \quad \textit{explanation} & ↑↑↑ & ↑ \\
        \quad \textit{clarification} & ↓↓↓ & ↓↓ \\
        \quad \textit{confirmation} & ↓↓↓ & ↓↓ \\
        \hline
        \textbf{Discourse regulators} & & \\
        \quad \textit{acknowledgment} & ↑↑↑ & ↓↓↓ \\
        \quad \textit{recipient} & ↓↓↓ & ↓ \\
        \quad \textit{theme} & ↓↓↓ & ↓↓↓ \\
        \quad \textit{enumeration} & ↓↓↓ & ↓↓↓ \\
        \quad \textit{counting} & ↓↓↓ & ↓↓↓ \\
        \quad \textit{inside comment} & ↓↓↓ & ↓↓↓ \\
        \hline
        \textbf{Prefaces} & & \\
        \quad \textit{reported speech} & ↓↓ & ↓↓↓ \\
        \quad \textit{opinion} & ↓↓↓ & ↓↓ \\
        \quad \textit{fact} & ↓↓↓ & ↓↓↓ \\
        \quad number & ↓↓↓ & ↓↓↓ \\
        \quad length & ↓↓↓ & ↓↓↓ \\
        \hline
        \textbf{Question types} & & \\
        \quad \textit{open} & ↓↓↓ & ↑ \\
        \quad \textit{polar} & ↓↓↓ & ↓↓↓ \\
        \quad \textit{closed-list} & ↓↓↓ & ↓↓ \\
        \hline
        \textbf{NLP features} & & \\
        \quad type-token ratio & ↑↑↑ & ↑↑↑ \\
        \quad Flesch-Kincaid & ↑↑↑ & ↑↑↑ \\
        \quad Dale-Chall & ↑↑↑ & ↑↑↑ \\
        \quad interjection count & ↑↑↑ & ↓↓↓ \\
        \quad word count & ↓↓↓ & ↓↓↓ \\
        \quad sentence count & ↓↓↓ & ↓↓↓ \\
        \quad filler word count & ↓↓↓ & ↓↓↓ \\
        \quad stopword count & ↓↓↓ & ↓↓↓ \\
        \quad NER (person) count & ↓↓↓ & ↓↓ \\
        \quad question count & ↓↓↓ & ↓↓ \\
        \quad assertion count & ↓↓↓ & ↓↓↓ \\
        \quad mean assertion length & ↓↓↓ & ↓↓↓ \\
    \end{tabular}
    \caption{Spearman correlations with perceived professionalism. Arrows indicate direction and strength: ↑ = p < .05, ↑↑ = p < .01, ↑↑↑ = p < .001 (↓ for negative).}
    \label{tab:all features}
\end{table}

\section{Use of AI Assistance in Writing}

AI language tools were utilized to enhance grammar and phrasing during the writing process. These tools were solely employed for linguistic refinement; all aspects of research design, analysis, and content development were independently conducted and verified by the research team.

\end{document}